\documentclass{article}
\usepackage{amssymb}
\usepackage[table]{xcolor}
\usepackage{multicol}
\usepackage[bookmarks=true]{hyperref}

\usepackage{algorithmicx}
\usepackage{algpseudocode}
\usepackage{amsmath}
\usepackage{multirow}
\usepackage{makecell}
\usepackage{bm}
\usepackage{booktabs}
\usepackage{graphicx}
\usepackage{bbding}
\usepackage{amssymb}
\usepackage{tikz}
\usepackage{appendix}
\usepackage[linesnumbered,ruled,vlined]{algorithm2e}

\usepackage[preprint]{corl_2025} 

\title{MoTo: A Zero-shot Plug-in Interaction-aware Navigation for General Mobile Manipulation}

\author{%
  Zhenyu Wu$^{1, 3*}$, Angyuan Ma$^{2,3,4*}$, Xiuwei Xu$^{2,3,4\ddagger}$, Hang Yin$^{2,3,4}$ \\
  \bf Yinan Liang$^{2,3,4}$, Ziwei Wang$^{5}$, Jiwen Lu$^{2,3,4}$, Haibin Yan$^{1\dagger}$ \\
  $^{1}$School of IEA, Beijing University of Posts and Telecommunications \\
  $^{2}$Department of Automation, Tsinghua University \\ 
  $^{3}$Beijing Key Laboratory of Embodied Intelligence Systems \\
  $^{4}$Beijing National Research Center for Information Science and Technology\\
  $^{5}$School of Electrical and Electronic Engineering, Nanyang Technological University \\
}

\makeatletter
\def\thanks#1{\protected@xdef\@thanks{\@thanks \protect\footnotetext{#1}}}
\makeatother

\thanks{$^{*}$Equal contribution. $^{\ddagger}$Project Leader. $^{\dagger}$Corresponding author: Haibin Yan (\href{mailto:eyanhaibin@bupt.edu.cn}{eyanhaibin@bupt.edu.cn}).}

\begin{document}
\maketitle

\vspace{-1cm}
\begin{abstract}
Mobile manipulation stands as a core challenge in robotics, enabling robots to assist humans across varied tasks and dynamic daily environments.
Conventional mobile manipulation approaches often struggle to generalize across different tasks and environments due to the lack of large-scale training.
However, recent advances in manipulation foundation models demonstrate impressive generalization capability on a wide range of fixed-base manipulation tasks, which are still limited to a fixed setting.
Therefore, we devise a plug-in module named MoTo, which can be combined with any off-the-shelf manipulation foundation model to empower them with mobile manipulation ability.
Specifically, we propose an interaction-aware navigation policy to generate robot docking points for generalized mobile manipulation. 
To enable zero-shot ability, we propose an interaction keypoints framework via vision-language models (VLM) under multi-view consistency for both target object and robotic arm following instructions, where fixed-base manipulation foundation models can be employed.
We further propose motion planning objectives for the mobile base and robot arm, which minimize the distance between the two keypoints and maintain the physical feasibility of trajectories.
In this way, MoTo guides the robot to move to the docking points where fixed-base manipulation can be successfully performed, and leverages VLM generation and trajectory optimization to achieve mobile manipulation in a zero-shot manner, without any requirement on mobile manipulation expert data.
Extensive experimental results on OVMM and real-world demonstrate that MoTo achieves success rates of 2.68\% and 16.67\% higher than the state-of-the-art mobile manipulation methods, respectively, without requiring additional training data. Our project homepage can be found at \href{https://gary3410.github.io/MoTo/}{here}.
\end{abstract}

\keywords{Mobile Manipulation, Vision Language Action Models} 


\section{Introduction}
	
Mobile manipulation allows robots to operate effectively in expansive environments, where solving complex tasks requires the joint coordination of the robotic arm and mobile base~\cite{xiong2024adaptive, yang2023harmonic, wu2025momanipvla}.
The rapid proliferation of intelligent robotic systems has created an urgent demand for mobile manipulation in diverse application domains, including household services~\cite {xiao2024robi, abbatematteo2024composable}, manufacturing~\cite{peng2024revolutionizing}, and logistics~\cite{vstibinger2021mobile}, where robots must autonomously execute cross-space manipulation tasks.
However, enabling robots to operate across diverse tasks in unstructured environments (e.g., human daily-life assistance) remains highly challenging.
Therefore, designing a general mobile manipulation framework for real-world deployments is desirable.

Existing mobile manipulation approaches~\cite{huang2023skill, yokoyama2023asc} propose end-to-end frameworks to jointly search the navigation and manipulation action spaces to improve performance. However, the end-to-end mobile manipulation policy models lack training on large-scale mobile manipulation datasets due to extremely expensive collection costs for mobile manipulation demonstration, which results in low generalization across tasks and environments.
On the other side, the emerging manipulation foundation models~\cite{li2024manipllm,kim2024openvla,liu2024rdt} have demonstrated high generalization capabilities across different manipulation tasks and diverse deployment scenes.
However, these models are not able to predict the movement of the base and thus are limited to fixed-base manipulation tasks.
A natural idea is to build a mobile manipulation framework by directly patching together the LLM or VLM with the manipulation foundation models. 
Conventional mobile manipulation frameworks employ navigation~\cite{yin2024sg, choi2024find} and fixed-base manipulation~\cite{fang2023anygrasp, ze2023gnfactor} modules separately to accomplish mobile manipulation tasks.
However, naive combining navigation and manipulation results in compounding errors since the large gap between the goals of navigation and manipulation~\cite{gu2022multi}.
Finding suitable navigation docking points for interaction remains a huge challenge in mobile manipulation tasks.

\begin{figure*}[t]
  \centering
    \includegraphics[width=0.98\linewidth]{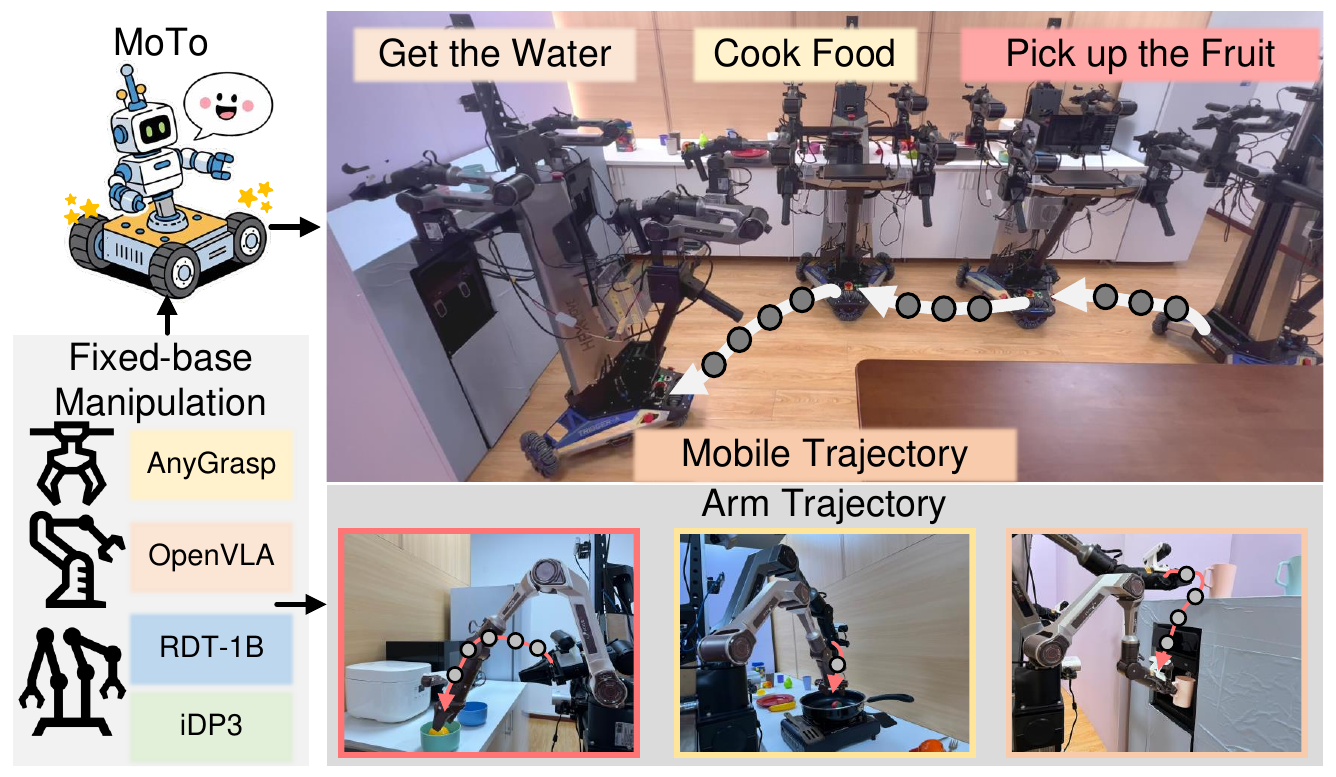}
    \vspace{-0.25cm}
   \caption{MoTo can be plugged into any fixed-base manipulation model and transferred to mobile manipulation tasks in a zero-shot manner, enabling generalized mobile manipulation.}
   \label{fig:teaser}
   \vspace{-0.7cm}
   \label{fig:compare}
\end{figure*}

In this paper, we propose to solve the problem of mobile manipulation with an interaction-aware navigation policy, namely \textbf{Mo}ve and \textbf{To}uch (MoTo). 
MoTo is a zero-shot plug-in that can be combined with any off-the-shelf fixed-base manipulation model and empower it with mobile manipulation ability. 
The proposed interaction-aware navigation policy generates suitable base docking points for the robot to successfully perform manipulation, which are inspired by the fact that the robot can perform subsequent fixed-base manipulation actions on the docking points where its arms can reach the target object interaction region.
We prompt the vision-language model (VLM) to generate task-related keypoints on the target object and robotic arm to enable zero-shot mobile manipulation. 
We can control the robot by solving an optimization problem that minimizes the distance between the two keypoints and considers several additional constraints, including collision avoidance, smoothness, and arm margin.
Figure \ref{fig:compare} illustrates the proposed approaches. MoTo efficiently transfers manipulation foundation models to diverse mobile manipulation tasks, where the mobile base and the robot arm coordinately perform actions with physically feasible trajectories.
Extensive experimental results on OVMM~\cite{yenamandra2023homerobot} and real-world demonstrate that MoTo achieves 2.68\% and 16.67\% higher success rates compared to state-of-the-art mobile manipulation techniques, respectively. 

\section{Related Works}
\textbf{Mobile Manipulation Frameworks:}
Existing mobile manipulation frameworks can be categorized into end-to-end and modular based on the model structure. 
The end-to-end approaches~\cite{brohan2022rt, fu2024mobile, qiu2024learning, zhang2025moma} directly map visual observations to the mobile manipulation action space.
M$^2$Diffuser~\cite{yan2024m2diffuser} guides the diffusion process with various energy terms to ensure that the trajectory of the mobile manipulation follows the physical constraints.
However, end-to-end approaches require extensive training data leading to high costs~\cite{mendonca2024continuously}.
The modular mobile manipulation framework~\cite{yenamandra2023homerobot, liu2024ok, rosen2023synthesizing} benefits from off-the-shelf navigation and manipulation frameworks exhibiting outstanding data efficiency. 
IALP~\cite{wang2025instruction} leverages LLM reasoning capability to generate manipulation actions in a closed-loop interactive manner. 
MoManipVLA~\cite{wu2025momanipvla} achieves data-efficient generalized mobile manipulation via optimization of the base waypoints to ensure that the VLA prediction trajectories are feasible.
AMR~\cite{meng2024aim} and LIMP~\cite{quartey2024verifiably} leverage language instructions to guide robots to target poses, demonstrating the potential of vision-language navigation for mobile manipulation.
However, naively combining navigation and manipulation complicates instructions and leads to compounding errors, as humans may not know optimal base positions for manipulation.

\textbf{Trajectory Optimization:}
Trajectory optimization is important for robots to perform precise manipulation actions.
Early research~\cite{kelly2017introduction, hewing2020learning, chang2016compositional} due to the hand‐crafted cost terms restricts deployment in diverse and complex environments.
To improve scalability, researchers focus on data-driven learnable methods. 
Contact-GraspNet~\cite{sundermeyer2021contact} and O2O-Afford~\cite{mo2022o2o} generate trajectories with learned grasping poses and object affordances, respectively.
Diffusion policy-based methods~\cite{chi2023diffusion, huang2023diffusion} learn collision-free trajectories directly from expert episodes and achieve promising performance.
More recently, foundation‑model–based frameworks like VoxPoser~\cite{huang2023voxposer} and ReKep~\cite{huang2024rekep} leverage pretrained priors to infer physical constraints, which significantly improve generalization.
Inspired by ReKep, we propose a multi-view voting strategy to generate scene-level interaction keypoints to fine-grain guide mobile manipulation trajectory generation.

\section{Problem Statement}


Our goal is to enable robots to perform long-horizon mobile manipulation tasks with strong generalization ability to unseen environments and goals. 
The task is described by a free-form language instruction $\mathcal{T}$. 
For example, ``I need to drink water", the robot should first pick up a bottle, put the bottle on a water dispenser, add water to the bottle, and place the bottle on a table near the human.

We consider a wheeled robot equipped with single or dual arms and RGB-D cameras. At each time step $t$, the robot receives RGB-D observations $(s_t^e, \{s_t^w\})$, and executes actions $(a_t^{base}, \{a_t^{arm}\})$, where $s_t^e$ is the observation from the exterior camera which captures the scene in front of the robot,  $\{s_t^w\}$ is the observation from wrist camera, $a_t^{base}$ is a 2-DoF robot base movement and $\{a_t^{arm}\}$ is 6-DoF arm pose and gripper status for each robotic arm.


Note that this work focuses on a general plug-in framework for mobile manipulation. 
We assume off-the-shelf fixed-base manipulation policies are available, which can complete low-level language instruction $\mathcal{T}_{fix}$.
Here, the task specified in $\mathcal{T}_{fix}$ must be completed with the fixed robotic base.

This work addresses the problem of determining base docking points from which the robotic arm can reach and manipulate the target object successfully.
\begin{equation}
\begin{aligned}
&\min \sum_{t=0}^{T}\mathcal{O}(\theta_t^{base}, \{\theta_t^{arm}\}) \\ 
&\text{s.t.} \quad \theta_t^{base} \in \mathbb{R}^{3}, \quad \theta_{t}^{arm} \in \mathbb{R} \\
\end{aligned}
\end{equation}
This equation represents the overall optimization objective $\mathcal{O}$ for the mobile manipulation trajectory.

With the fast development of manipulation foundation models~\cite{huang2023voxposer,kim2024openvla,liu2024rdt,huang2024rekep}, we believe this assumption is reasonable and feasible. And what we should accomplish is a general plug-in that can be combined with any off-the-shelf fixed-base manipulation model and empower it with mobile manipulation ability.

\begin{figure*}[t]
  \centering
    \includegraphics[width=1.0\linewidth]{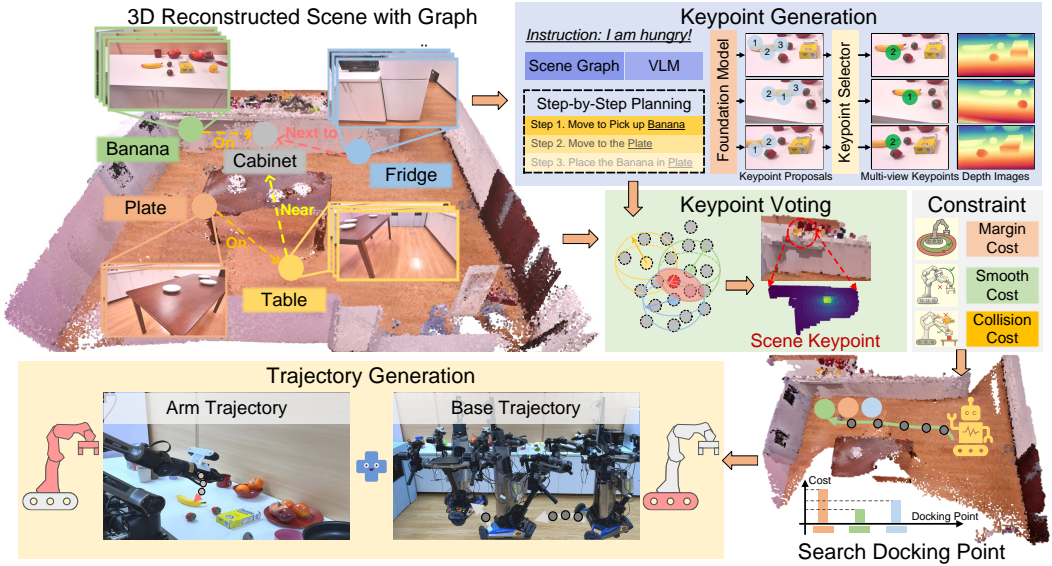}

   \caption{The pipeline of MoTo. Based on robot scanning RGB-D observation to get 3D scene point clouds and graphs, we utilize VLM and multi-view consistency voting to get interaction keypoints, and generate mobile manipulation trajectories via proposed cost constraint optimization.}
   \label{fig:pipeline}
   \vspace{-0.5cm}
\end{figure*}

\section{Approach}
\subsection{Interaction-aware Navigation}
Most mobile manipulation tasks can be decomposed into subtasks, each involving a single target object.
For example, $\mathcal{T}$: "I want to drink milk." can be divided into: (1) pick up a \textbf{bottle}; (2) place the bottle on the \textbf{table}; (3) pick up the \textbf{milk}, etc.
Each subtask has one target object~(in \textbf{bold}).
Two objects are involved in subtask (2): one in the robot's gripper and another to be interacted with, the latter being defined as the target object.
We assume each subtask can be completed by first navigating and then manipulating with a fixed base\footnote{Navigation is optional; if the robot can operate the target object with its arm without moving the base, this step is unnecessary. We do not consider tasks requiring simultaneous base and arm movement, such as opening a door with a rotating handle.}.
Since fixed-base manipulation policies are available, the main challenge for our work is how and where to move, i.e., the last-mile challenge.

Determining how and where to move is a non-trivial problem that cannot be solved by a pure navigation policy.
A robot must stop at a docking point for feasible manipulation, where the target object is at a suitable orientation and distance with no significant obstacles.
Since high-quality mobile manipulation data is limited, we tackle this problem in a zero-shot manner by introducing interaction-aware navigation.
We observe that a subtask becomes easy with a fixed base if (1) the robot can reach the interaction-relevant keypoints of the target object and (2) the arm retains sufficient motion margin.
The keypoints must be part-level. 
Imagine the robot is going to open a fridge. 
The robot cannot open without moving the base if it reaches the back of the fridge.
Similarly, the robot must choose context-specific interaction points.
If the gripper is empty, the end-effector can be used; if holding a tool, such as a broom, the relevant part (e.g., broom tip) must engage with the environment (e.g., contacting trash on the floor).
Based on this insight, interaction-aware navigation decomposes the movement problem into two subproblems: (1) identifying the target keypoint (TK) on the object and the arm keypoint (AK) on the manipulator, and (2) controlling the robot to move until TK and AK are aligned for interaction.

\textbf{Approach Overview.} To accurately ground instruction $\mathcal{T}$ to objects in the scene, we first let the robot scan the whole scene and reconstruct a 3D point cloud $P\in \mathbb{R}^{N\times 3}$ along with a 3D scene graph $\mathcal{G}=(\mathcal{N}, \mathcal{E})$. 
Each object node in $\mathcal{N}$ stores the 3D coordinate, 3D instance mask and several object-centric images $\{I_1^k,...,I_m^k\}$ of an object $o_k$. The edge in $\mathcal{E}$ indicates the spatial relationship between two objects, such as ``inside" and ``on top of".
With $\mathcal{G}$, we can prompt an LLM to perform task planning based on the existing objects and their relationship:
\begin{equation}
    \{(\mathcal{T}_1,o_1), ..., (\mathcal{T}_n,o_n)\}={\rm LLM}(\mathcal{T}, \mathcal{G})
\end{equation}
where $\mathcal{T}_k$ is the $k$-th subtask. $o_k\in \mathcal{N}$ is the corresponding target object, which should be explicitly referred to in the language description of $\mathcal{T}_k$.
We can also query the object-centric image set $\{I_1^k,...,I_m^k\}$ from $\mathcal{G}$ with $o_k$.
Based on these, we leverage vision-language models (VLM) to obtain target keypoint $\mathcal{P}_k^T$ and arm keypoint $\mathcal{P}_k^A$ in the world coordinate system:
\begin{equation}\label{generation}
\begin{aligned}
    \mathcal{P}_k^T&=\mathcal{V}({\rm VLM}(\mathcal{T}_k, \{I_1^k,...,I_m^k\}), P)\\
    &\mathcal{P}_{k,t}^A=\Gamma({\rm VLM}(\mathcal{T}_k, \{s_t^w\}), E_t)
\end{aligned}
\end{equation}
where we consider the scene point clouds $P$ in the world coordinate system. 
$\Gamma$ represents the mapping function from the robot arm coordinate system to the camera coordinate system.
${\rm VLM}(\mathcal{T}_k, \{I_1^k,...,I_m^k\})$ generates target keypoint proposals in different images, which are then aggregated with a voting module $\mathcal{V}$.
${\rm VLM}(\mathcal{T}_k, \{s_t^w\})$ first generates arm keypoint in the robot's coordinate system, which is then transformed to the world coordinate system via the mapping function $\Gamma$ using the robot-to-world transformation $E_t$ acquired with off-the-shelf SLAM method.
Note that $\mathcal{P}_{k,t}^A$ is a function of the position and pose of the robot base and arm. Therefore, the robot's action $(a_t^{base}, \{a_t^{arm}\})$ can be solved as an optimization problem, which aims to minimize the distance between TK and AK as well as following several constraints:
\begin{equation}\label{optimization}
    \mathop{\rm argmin}_{a_{1:T}^{base}, \{a_{1:T}^{arm}\}} \sum_{t=1}^{T}||\mathcal{P}_k^T-\mathcal{P}_{k,t}^A(\theta_t^{base}, \{\theta_t^{arm}\})||+\mathcal{C}_t
\end{equation}
where $\theta_t^{base}$ and $\{\theta_t^{arm}\}$ represent the position and pose of robot base and arm respectively. $\mathcal{C}_t$ is the additional constraints.
We will detail Eq (\ref{generation}) and (\ref{optimization}) in Section \ref{sec:gen} and \ref{sec:opt} respectively.

\subsection{VLM-based Keypoint Generation}\label{sec:gen}

Since high-quality data on instruction-affordance pairs is limited, training a model to generate reasonable TK and AK according to text description $\mathcal{T}_k$ is very challenging. 
However, we notice that although current models are unable to directly generate task-related keypoints on an object, VLMs demonstrate great ability to select a proper keypoint among a set of candidates given the task description.
Therefore, we propose a two-stage VLM-based method to generate keypoints for an image, which is divided into keypoint proposal stage and keypoint selection stage. 

\textbf{Keypoint Proposal.}
In this stage, we exploit visual cues in the given image to generate several keypoint proposals. Without any language information, we aim to find out all representative actionable points in an image as proposals. Note that this is more challenging than part segmentation, because actionable points may not have significant color or texture differences. For example, to close a laptop we should push the back of the screen, to lift a table we should hold both centers of its opposite edges. These points to be interacted with do not belong to a specific part. Current segmentation models can only segment a laptop into screen and keyboard, and a table into surface and legs, which cannot provide detailed, actionable locations. Inspired by ReKep~\cite{huang2024rekep}, we adopt DINOv2~\cite{oquab2023dinov2} to extract pixel-wise representation of the image, where semantically meaningful regions show larger activations. Then we apply SAM~\cite{kirillov2023segment} to extract fine-grained part-level masks. Finally, in each segmented mask, we apply clustering on the DINOv2 features normalized with depth information, and select the center of each cluster as the keypoint proposal\footnote{If no keypoints are provided, the end-effector’s proprioceptive position is used as the default. }.

\textbf{Keypoint Selection.}
Given the extracted keypoint proposals, we index and plot them on the image. We then prompt VLM with this modified image as well as the task description $\mathcal{T}_k$. VLM will select the most suitable keypoint from the proposals.


We adopt a two-stage pipeline to obtain AK. Firstly, extracting the wrist keypoint from the RGB-D observation ${s_t^w}$, then projecting it to 3D space using $E_t$. If the gripper is empty, we skip this process and directly set the end-effector’s 3D position as AK.
For dual-arm robots, both arms may output keypoints, which confuses the following optimization process. To address this, we adopt VLM as a binary classifier to select a specific arm for the current subtask, ensuring only one AK contacts TK. The VLM is prompted with $\mathcal{T}_k$, the Euclidean distances from each end-effector to the target, and the gripper contents (“empty” or object category) to determine the appropriate arm.

\textbf{Multi-view Voting for TK.}
Different from AK, TK should be generated on the target object, which may be far away from the robot and is harder to be backprojected accurately. So we further propose a voting module $\mathcal{V}$ to aggregate multi-view predictions into the final TK in 3D space.

To enable robust voting, we modify the keypoint selection prompt to let VLM select the top $V$ most relevant keypoint proposals. In this way, we generate $V\times m$ keypoints for the image set $\{I_1^k,...,I_m^k\}$ of target object. These keypoints are projected to 3D space with corresponding depth images and camera parameters, which are denoted as $P_K\in\mathbb{R}^{(V\times m)\times 3}$.
We then utilize $P_K$ to vote for the points $P$ in the scene:
\begin{equation}
    \mathcal{V}(P_K,P)=\mathop{\rm argmax}_{i}\sum_{j=1}^{V\times m}{\mathbb{I}(||P(i)-P_K(j)||_2-\tau)}
\end{equation}
where $\mathbb{I}$ is an indicator function that converts negative values to 1 and others to 0, and $\tau$ is a Euclidean distance threshold. We use the voting result to index $P$ and obtain the 3D point of TK.

\subsection{Keypoint-guided Optimization}\label{sec:opt}

With the generated TK and AK, now we solve an optimization problem to make them touch with each other to control the robot moving to a right point.
Since the robot-to-world transformation $E_k$ depends on the position and pose of robot base and arm, the arm keypoint $\mathcal{P}_{k,t}^A$ can be written as a function of $\theta_t^{base}$ and $\{\theta_t^{arm}\}$.
Note that we have:
\begin{equation}
    a_t^{base}=\Delta\theta_{t}^{base},\ \ \{a_t^{arm}\}=\Delta\{\theta_{t}^{arm}\}
\end{equation}
Therefore by formulating $\mathcal{C}_t$ as the function of $\theta_t^{base}$ and $\{\theta_t^{arm}\}$, we can solve Eq (\ref{optimization}) to obtain the optimal trajectories of robot base and arm, from which the actions can be easily derived with Pinocchio IK solver~\cite{carpentier2019pinocchio}.

In addition to making TK and AK touch each other, we also need to design constraints $\mathcal{C}_t$ to ensure the smooth execution of the control, including collision avoidance, control smoothness, and the margin of the robotic arm. We formulate three cost functions to achieve this:
\begin{equation}
    \mathcal{C}_t(\theta_t^{base}, \{\theta_t^{arm}\})=\mathcal{F}_t^c+\mathcal{F}_t^s+\mathcal{F}_t^m
\end{equation}

\textbf{Collision Cost.}
The robot is required to avoid any collision with objects in the scene to ensure safety during the mobile manipulation process. We uniformly sample $N_q$ query points on the robot surface $\Omega$ to evaluate the collision cost:
\begin{equation}
    \mathcal{F}_t^c=\sum_{j=1}^{N_q} \max(0, ~\epsilon_0-\mathcal{D}(q_t^j,P)), \quad q_t^j \in \Omega(\theta_{t}^{base}, \{\theta_t^{arm}\}) 
\end{equation}where $q_j^t$ means the $j$-th query point, $\epsilon_0$ is a hyperparameter that controls the safety margin for collision avoidance. $\mathcal{D}(q_j^t, P)$ means the distance between $q_j^t$ and the scene point clouds $P$. 
We expect the distance between the robot surface and the scene can be maximized if it is within the safety margin. Otherwise, the collision cost will not influence the control of the robot.

\textbf{Smoothness Cost.}
This cost keeps continuous and smooth changes for joint angles of the robot arm and translation and rotation of the base, where sudden changes are avoided to maintain the safety of the motor and the embodiment. 
We define trajectory smoothness as the difference of joint angles $\bm{\theta}^{t}$ of the arm and proprioception of the base between consecutive poses in the candidate trajectory:
\begin{equation}
    \mathcal{F}_t^s=||\theta_{t+1}^{base}-\theta_{t}^{base}||_2+\sum||\theta_{t+1}^{arm}-\theta_{t}^{arm}||_2
\end{equation}
We leverage the joint angle solved via IK for smoothness constraint instead of the proprioception of the arm, because small changes in the arm pose cannot guarantee slight difference in joint angles.

\textbf{Margin Cost.}
When AK and TK make contact, a large margin is required for the robotic arm to enable subsequent fixed-base manipulation. We define the horizontal distance between the end-effector and the robot base center as the arm radius $r$, bounded by constants $r_{min}$ and $r_{max}$.
Given the robot’s design, $r$ is a function of the arm pose $r=g(\theta_{t}^{arm})$.
The margin cost is defined as:
\begin{equation}
    \mathcal{F}_t^m=||\frac{r_{min}+r_{max}}{2}-g(\theta_{t}^{arm})||_2
\end{equation}
In this way, the robotic arm will not be too curved or too extended, ensuring that it has a large margin to operate.

In the process of optimization solving, the decision variables include the robotic arm joint angle $\{\theta_{t}^{arm}\}$ and base pose $\theta_{t}^{base}$. 
Specifically, we generate initial solution proposals for the next time step from the search space and calculate the cost relative to the optimization objective for each proposal.
Using the Dual Annealing~\cite{xiang1997generalized} algorithm, we iteratively update the initial solutions based on the change in cost, continuing this process until the cost is reduced below a predefined threshold.

\section{Experiment}



\begin{table*}[t]
\caption{Comparison results on the OVMM benchmark. Partial success rates indicate the execution of each stage, conditioned on the success of the preceding one. “RL” and “Heuristic” refer to object placement methods based on RL and heuristics, respectively. The RL method is used by default.}
\centering
\small
\setlength{\tabcolsep}{6.2pt}
\begin{tabular}{@{}lccccccc@{}}
\toprule
\multirow{2}{*}{\textbf{Method}}
& & \multicolumn{3}{c}{\textbf{Partial Success Rates}} &\multirow{2}{*}{\textbf{\makecell{Overall
\\ SR}}} &\multirow{2}{*}{\textbf{\makecell{Average
\\ SR}}} &\multirow{2}{*}{\textbf{Step}} \\
\cmidrule{3-5} 
  & & FindObj~($\uparrow$) & Pick~($\uparrow$) & FindRec~($\uparrow$)  \\
 \midrule
Home-Robot (RL)&  &66.60\%  & 61.10\% &  50.90\% & 14.80\% & 48.30\% & 1132.5  \\
Home-Robot (Heuristic) &  &65.40\%  &54.80\% &43.70\% &7.30\% &42.80\% & 1009.8  \\
UniTeam~\cite{melnik2023uniteam} &  &66.13\% &62.65\%  &54.69\% &  17.96\% & 50.36\% & 1027.7  \\
\midrule
 Home-Robot w/ L3MVN   & & 68.10\%   & 60.14\%  & 57.46\%   & 15.28\%    & 50.25\%   & 1084.3 \\
 OpenVLA w/ L3MVN & & 68.28\%   & 60.68\%    & 58.00\%   & 16.09\%   & 50.76\%  & 1191.5 \\
 \midrule
 \rowcolor{gray!20}
 Home-Robot w/ MoTo~(Ours) & &65.24\% &60.23\% &50.40\% &18.32\% &48.55\% &1152.2 \\
 \rowcolor{gray!20}
 OpenVLA w/ MoTo~(Ours) & &66.67\% &60.95\% &49.87\% &20.64\% &49.53\% &1195.1\\

 \midrule
\end{tabular}
\label{table_1:OVMM}
\vspace{-0.5cm}
\end{table*}

\begin{table*}[t]
\caption{Ablation experiments for optimization cost terms and keypoint generation variants. }
\centering
\small
\setlength{\tabcolsep}{9pt}
\begin{tabular}{ccccccc}
\toprule
\multicolumn{2}{c}{\multirow{2}{*}{\textbf{Method}}}                         & \multicolumn{3}{c}{\textbf{Partial   Success Rates}} & \multirow{2}{*}{\begin{tabular}[c]{@{}c@{}}\textbf{Overall}\\      \textbf{SR}\end{tabular}} & \multirow{2}{*}{\begin{tabular}[c]{@{}c@{}}\textbf{Average}\\      \textbf{SR}\end{tabular}} \\
\cmidrule{3-5} 
\multicolumn{2}{c}{}     & FindObj~($\uparrow$)       & Pick~($\uparrow$)         & FindRec~($\uparrow$)      &       &     \\
 \midrule
\multirow{3}{*}{Cost}& w/o   Collision  & 66.93\% & 60.95\%  & 49.24\%  &18.50\%  & 48.91\%    \\
& w/o Smoothness   & 66.76\%   & 61.48\%   & 49.60\%    & 19.75\%     & 49.40\%  \\
& w/o   Margin    &65.95\%        &60.77\%       & 50.04\%      &17.87\%        &48.66\%         \\
 \midrule
\multirow{4}{*}{Keypoint}  & SAM+CLIP   & 66.31\%   & 58.36\%   & 50.58\%   & 20.29\%     & 48.88\%   \\
& SAM+ViT      & 65.95\%   & 57.19\%   & 49.87\%    & 20.46\%     & 48.37\%    \\
& w/o Fusion   & 66.13\%  & 56.21\%  & 49.96\%   & 15.19\%    & 46.87\%    \\
& Single View  & 66.49\%  & 58.27\%  & 50.04\%   & 17.61\%    & 48.19\%     \\
\toprule

\end{tabular}
\label{table_1:ab}
\vspace{-0.4cm}
\end{table*}

\subsection{Comparison with State-of-the-art Methods}
Table~\ref{table_1:OVMM} demonstrates the performance of MoTo on the OVMM~\cite{yenamandra2023homerobot} validation set compared to the baseline, decomposing it into four sequential stages: finding the target~(FindObj), grasping~(Pick), finding the container~(FindRec), and placing~(Overall SR). 
Our training and testing methods fully comply with the OVMM baseline settings~(Home-Robot).
The implementation details are in Appendix~\ref{sec: A.1}.~The similar success rates in stages FindObj and Pick are due to MoTo's focus on interaction-aware navigation, which is invoked only after finding a container without exploring it.
Home-Robot w/ MoTo compared to Home-Robot~(RL) in Table~\ref{table_1:OVMM} illustrates MoTo’s interaction-aware navigation yields an overall success rate improvement of 3.52\%.~With a stronger manipulation foundation~(e.g., OpenVLA~\cite{kim2024openvla}), MoTo’s success rate can increase by an additional 2.32\%.
We also compare the performance of MoTo with that of the conventional navigation framework~(L3MVN~\cite{yu2023l3mvn}) on the mobile manipulation task. 
Since the conventional framework only ensures that the robot reaches the target nearby and ignores the feasibility of subsequent manipulations, which results in a marginal success rate gain~(Overall SR only increased by 0.48\%) despite a 6.65\% performance improvement in the FindRec stage.

\begin{figure}[t]
  \centering
  \includegraphics[width=1.0\linewidth]{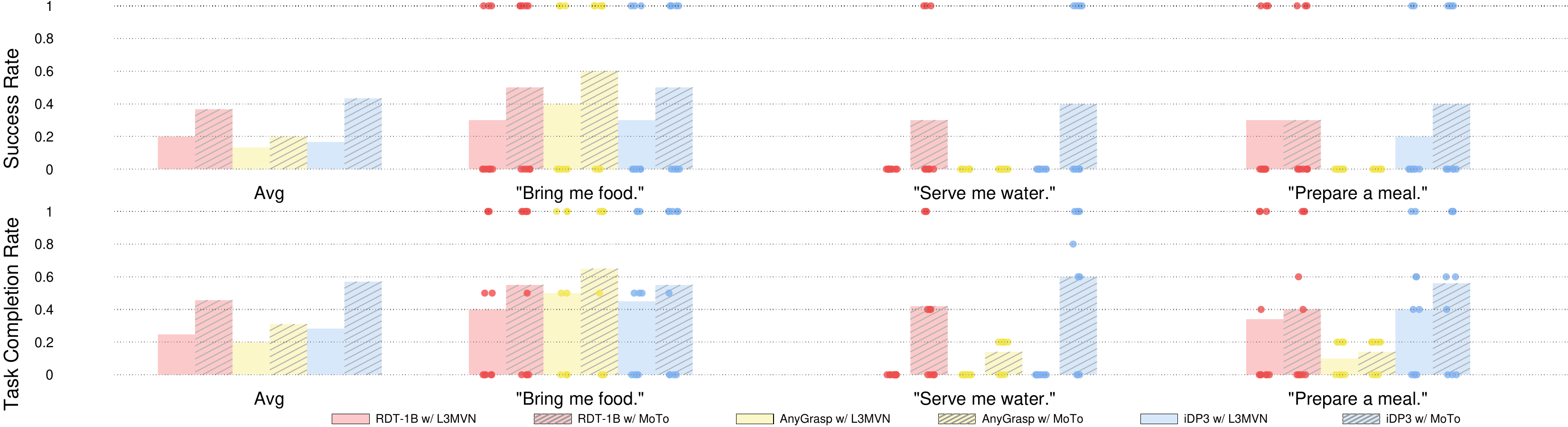}
  \caption{Real-world experimental results. All methods are run 10 times on the three types of mobile manipulation tasks, where the dots represent the performance of each test~(Best view in color).}
  \label{fig:real world}
  \vspace{-0.5cm}
\end{figure}

\subsection{Ablation Study}
Table \ref{table_1:ab} reports a systematic ablation of our full MoTo pipeline in OVMM, isolating the contribution of each optimization cost term and keypoint design choice, where w/o Fusion denotes that no voting is applied on the generated multi-view keypoints, and Single View means only the view with the most target pixels is utilized.
Experimental results demonstrate that each constraint cost in MoTo can improve performance.
Since the OVMM involves only the pick and place task, the margin cost is more important than the others and determines whether the arm can reach the target position or not.
We further investigated variants of the keypoint extraction and fusion pipeline. 
The inconsistency of multi-view keypoints in the ``w/o Fusion" setting results in a serious performance drop (2.42\% lower success rate compared to Single View), because the cluttered keypoints fail to provide effective guidance.
Wrong interaction keypoints will result in the robot moving elsewhere due to physical constraints that prevent it from performing fixed-base manipulation actions.
The choice of visual foundation model for keypoint extraction has little impact on MoTo; replacing DINOv2 with CLIP or ViT still yields high performance.


\subsection{Real World Experiments}
Due to the sim-to-real gap, the OVMM baseline cannot be deployed directly in the real world. Therefore, to further verify the plug-in scalability of our proposed MoTo, we deployed diverse fixed-base manipulation models (AnyGrasp~\cite{fang2023anygrasp}, iDP3~\cite{ze2024generalizable}, RDT-1B~\cite{liu2024rdt}) covering heuristics, diffusion policy, and foundation models in the real world, as illustrated in Figure \ref{fig:real world}. 
Task completion rate indicates the number of subtasks completed, and the success rate is 1 only if all subtasks are completed.
The deployment and evaluation details are in Appendix A.2 and B.
Compared to fixed-base manipulation with conventional navigation apportion, MoTo achieves performance improvement in all test tasks, which illustrates the efficiency of MoTo in mobile manipulation and points out the powerful scalability of the plug-in. 
Since AnyGrasp can only perform the grasping action, its performance is low on complex interaction tasks such as “I am thirsty”.~Interestingly, iDP3 and AnyGrasp outperform the foundation model RDT-1B in mobile manipulation, largely due to their superior viewpoint generalization from 3D point cloud observations. This capability helps address viewpoint changes caused by diverse docking points. These results suggest that exploring 3D egocentric manipulation policies is a promising direction for mobile manipulation.

\section{Conclusion}
In this paper, we present a plug-in interaction-aware navigation module named MoTo, which can efficiently transfer existing fixed-base manipulation models to mobile manipulation tasks in a zero-shot manner. Specifically, we utilize vision foundation models to generate interaction keypoints for the manipulation task, and leverage multi-view consistency to fuse the keypoints to accurately provide guides for robot base and arm motion planning. 
We further propose motion planning objectives based on arm and goal keypoints and cost constraints to improve the physical feasibility of trajectories. 
Extensive experiments in the real world and simulators verify the effectiveness of MoTo and scalability for arbitrary manipulation model compatibility.

\section{Limitation}
Although the proposed MoTo can be plug-and-play with any off-the-shelf fixed-base manipulation model for efficient mobile manipulation, real-world deployments still have some limitations. First, MoTo lacks the ability to reconstruct the scene online, which prevents deployment in dynamic environments. Second, MoTo relies on interaction-aware navigation to generate docking points for fixed-base manipulation models but does not incorporate sufficient whole-body control policies, making it struggle with complex tasks such as door opening. Third, MoTo makes a strong assumption that all relevant objects are fully observable within the agent’s current view and at sufficient resolution to allow reliable keypoint extraction (e.g., ‘we first let the agent scan the whole scene’). This assumption limits robustness in partially observable or cluttered settings.

In future work, we plan to equip MoTo with incremental online scene reconstruction and perception algorithms to handle dynamic and partially observable environments. Additionally, we aim to integrate cost constraints directly into policy generation (e.g., during the denoising step of Diffusion Policy) to produce fine-grained whole-body control trajectories, thereby enabling more challenging mobile manipulation tasks.

\acknowledgments{This work was supported in part by the National Natural Science Foundation of China under Grant 62376032 and 624B2076, in part by the Beijing Natural Science Foundation under Grant No. L257008, and in part by the MoE AcRF Tier 1 Seed Grant RS17/24 and NTU EEE Ignition Research Grant 024920-00001.}


\bibliography{example}  

\clearpage

\appendix


\section{Implementation Details}
\subsection{Simulator Experiment}
\label{sec: A.1}
The OVMM benchmark consists of 60 extensive indoor scenes and contains more than 18k 3D models of everyday objects.OVMM utilizes Hello Robot as an agent to perform the “Move a target object from container A to container B” mobile manipulation task. We utilize an OVMM-heuristic baseline to collect manipulation expert trajectories that include robot proprioception, action, and visual observations to fine-tune off-the-shelf manipulation foundation models. 
The simulator experiments are training and testing on 8 RTX 3090 GPUs.
MoTo leverages ConceptGraph~\cite{gu2024conceptgraphs} to build 3D scene point clouds and scene graphs from RGB-D inputs.
The object nodes and edges in the scene graph are converted into a structured description to query the VLM for target identification.

\begin{figure}[h]
  \centering
  \setlength{\abovecaptionskip}{-0.1cm}
  \includegraphics[width=1.0\linewidth]{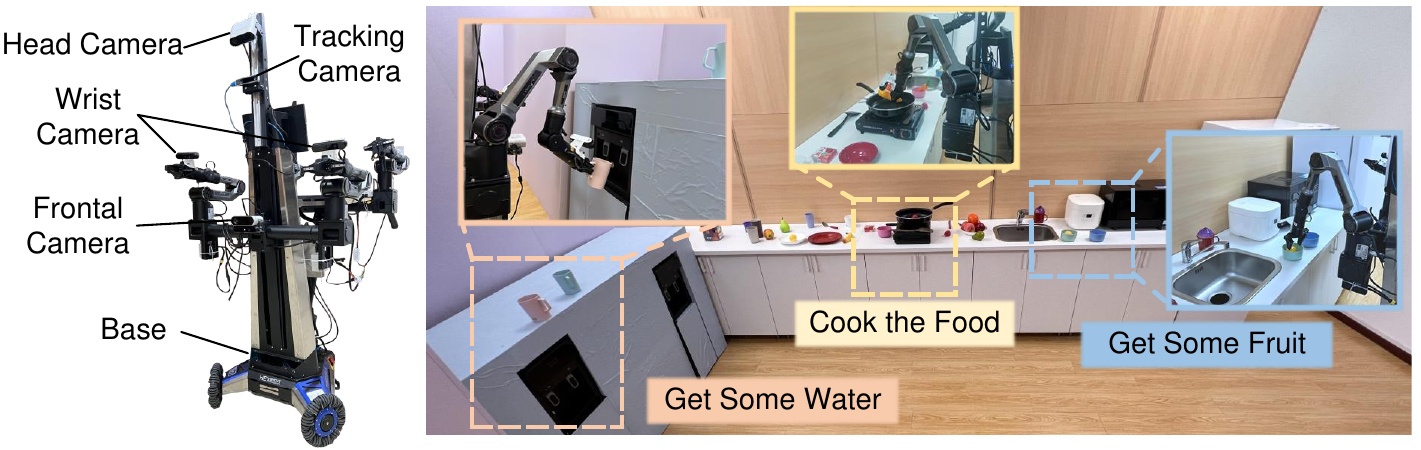}
  \caption{Real-world experimental platforms and deployment environments.}
  \label{fig:env_vis}
  \vspace{-0.5cm}
\end{figure}

\subsection{Real World Experiment}
For the real-world experiments, we use HEXMOVE as the base and two PiPER robotic arms to build a dual-arm mobile manipulation robot, which is equipped with a Femto Bolt RGB-D sensor as the head camera to acquire high-quality scene point clouds, and two Gemini 336L RGB-D sensors as the wrist cameras to assist in the manipulation task execution. Meanwhile, an Intel tracking camera T265 is employed to acquire the robotic camera poses to reconstruct the deployment scene. We recognize scene target objects and task planning using GroundingSAM~\cite{ren2024grounded} and GPT-4o, respectively, which ensure the zero-shot generalization ability of the overall framework.
MoTo also leverages ConceptGraph~\cite{gu2024conceptgraphs} to build 3D scene point clouds and scene graphs from RGB-D inputs.
Figure \ref{fig:env_vis} further demonstrates the real-world experiment platform and environment.
The real-world experiments are all performed on a single RTX 4060 GPU.

\subsection{Training Details}
In the OVMM simulator, we collect expert demonstration data for pick-and-place during mobile manipulation with heuristic baselines.
To better fine-tune OpenVLA~\cite{kim2024openvla} to mitigate cross-robot ontology differences, we collected a total of 20k data and fine-tuned 10k epoch on 8 RTX 3090 GPUs using the LoRA strategy.
We ensured the diversity of viewpoints in the expert trajectories during the collection process in order to achieve a higher viewpoint generalization.

For both RDT-1B~\cite{liu2024rdt} and iDP3~\cite{ze2024generalizable} manipulation policies, we collected fine-tuning data in a realistic kitchen environment, spanning three long-horizon tasks—\emph{Arrange various food items}, \emph{Dispense hot water from a water dispenser and mix with cold water}, and \emph{Cook food}—each decomposed into 2-5 subtasks (e.g., “pick up ingredient,” “align container,” “adjust water temperature”) with 50 expert demonstrations per task.

\textbf{RDT-1B:} The RDT-1B policy uses dual-arm 6-DOF joint positions, gripper open/close angles, and synchronized RGB streams (640×480 @ 30 Hz) from three cameras—one frontal and two mounted on the left and right grippers—as inputs. We fine-tuned the model for 150{,}000 gradient steps on 8 NVIDIA RTX 4090 GPUs (total batch size 128) using the AdamW optimizer (learning rate $1\times10^{-4}$, weight decay $1\times10^{-2}$), following the default RDT-1B configuration.

\textbf{iDP3:} The iDP3 policy takes dual-arm 6-DOF joint positions, gripper angles, and point-cloud frames (640×480 depth → 3D XYZRGB) from a frontal Orbbec Femto Bolt (20 Hz) as inputs. We reduced both the agent-state vector and action vector to 14 dimensions (7 per arm) to match our dual-arm kinematics. Training was performed for 3{,}000 epochs on a single RTX 4090 GPU (batch size 64) using Adam (learning rate $1\times10^{-4}$, weight decay $1\times10^{-6}$). Validation performance plateaued after approximately 2{,}500 epochs, and we selected the checkpoint with the lowest validation loss.

\section{Real World Tasks and Metrics}
\subsection{Task Description}
In real-world experiments, we have tested three types of task instructions to verify that MoTo can fully utilize the off-the-shelf fixed manipulation model to accomplish the task requirements.
Specifically, each task corresponds to multiple instructions, and the robot needs to reason about the best interaction goal as well as action planning based on the scene information.
Details of the test tasks are demonstrated in Table~\ref{tb:instruction}.

\begin{table}[t]
\small
\setlength{\tabcolsep}{2.5pt}
\caption{Real-world task instructions. The target is the object to be interacted with, and the subgoal is the condition that needs to be completed. }
\begin{tabular}{c|l|c|l}
\toprule
Type & \multicolumn{1}{c|}{Instruction} & Target & \multicolumn{1}{c}{Subgoal} \\ 
\midrule
\multirow{5}{*}{Bring me food} & Give me a   banana & Banana & \multirow{5}{*}{\begin{tabular}[c]{@{}l@{}}1.   Grasp the target\\      2. Place the target in the plate\end{tabular}} \\ \cline{3-3}
 & Get some food   for me & \multirow{3}{*}{\begin{tabular}[c]{@{}c@{}}Strawberry, \\ Banana or Lemon\end{tabular}} &  \\
 & I want to eat some food &  &  \\
 & I am hungry &  &  \\ \cline{3-3}
 & I want to eat a strawberry & Strawberry &  \\ \midrule
\multirow{5}{*}{Serve me water} & I want a cup   of water & Cup & \multirow{5}{*}{\begin{tabular}[c]{@{}l@{}}1.   Grasp the target\\      2. Fetch hot water  \\ 3. Grasp target filled with cold water\\4. Mix hot and cold water \\ 5. Put down the targets \end{tabular}} \\ \cline{3-3}
 & I want to   drink water & \multirow{3}{*}{\begin{tabular}[c]{@{}c@{}}Cup, \\ Mug or Bowl\end{tabular}} &  \\
 & I need drink &  &  \\
 & Give me some water, please! &  &  \\ 
 & & & \\ \midrule
\multirow{5}{*}{Prepare a meal} & Prepare a launch & \multirow{4}{*}{\begin{tabular}[c]{@{}c@{}}Banana, Strawberry \\ or Corn \end{tabular}} & \multirow{5}{*}{\begin{tabular}[c]{@{}l@{}}1.   Grasp the target\\      2. Put the target in the Pan\\     3. Grasp the bowl \\ 4. Grasp the cooked target\\      5. Put the target in the Bowl\end{tabular}} \\
 & Make some   salad &  &  \\
 & Warm up the   food &  &  \\
 & Cook for a   meal &  &  \\ 
  & & & \\  \toprule
\end{tabular}
\label{tb:instruction}
\vspace{-0.5cm}
\end{table}

\subsection{Metrics}
We follow the metrics of~\cite{shridhar2020alfred} and use the success rate and task completion rate to measure real-world mobile manipulation performance. 
For each instruction, the robot must complete the corresponding subtasks in Table~\ref{tb:instruction} for the manipulation to succeed. 
Task completion rate is the ratio of completed subgoals to those necessary to complete a task, which reflects the progress of the robot in completing the task. For example, in the “Serve me water” task, if the robot only successfully grasps and places the target (e.g., a cup or bowl), but there is no water in the target, the task completion rate is 0.4.

\section{Optimization Algorithm}
Navigation policy is one of the fundamental challenges of mobile manipulation, and it must ensure that the generated base docking points are feasible for subsequent actions. 
Conventional navigation methods only provide coarse docking point proposals that cannot be effectively applied to mobile manipulation. 
For this reason, we propose the optimization objective of Eq. 1 with the cost constraint of Eq. 3 to ensure the feasibility of the docking point selection utilizing an optimization search approach. 
Specifically, it is very difficult to search for mobile manipulation trajectories directly in large-scale scenes, and we use an existing navigation algorithm to move closer to the target object to reduce the optimized search space, searching for the optimal arm joint angle $\{\theta_{t}^{arm}\}$ and base pose $\theta_{t}^{base}$ iteratively with the Double Annealing Algorithm, as illustrated in Algorithm 1.

\begin{algorithm}[H]
\caption{Dual Annealing-Based Trajectory Optimization for Mobile Manipulation}
\KwIn{Initial base pose $\theta_0^{base}$, arm joint configuration $\{ \theta_0^{arm}\}$, time horizon $T$, cost function $\mathcal{O}$, temperature schedule $T_{anneal}$}
\KwOut{Optimized trajectory $\{\theta_t^{base}, \{\theta_t^{arm}\}\}_{t=0}^{T}$}

\For{$t = 0$ \KwTo $T$}{
    Generate candidate proposals $\{(\hat{\theta}_t^{base}, \{\hat{\theta}_t^{arm}\})^k\}_{k=1}^{K}$\;
    Evaluate cost for each proposal: $C_k = \mathcal{O}(\hat{\theta}_t^{base}, \{\hat{\theta}_t^{arm}\})$ \tcp*{Use Eq.~(4)}
    Select best candidate $(\theta_t^{base}, \{\theta_t^{arm}\}) = \arg\min_k C_k$\;

    \Repeat{cost improvement $< \varepsilon$ or max iterations reached}{
        Propose new candidate $(\hat{\theta}_t^{base}, \{\hat{\theta}_t^{arm}\})$ using Dual Annealing schedule $T_{anneal}$\;
        Compute cost $C_{new} = \mathcal{O}(\hat{\theta}_t^{base}, \{\hat{\theta}_t^{arm}\})$\;
        \If{acceptance criterion satisfied}{
            Update current solution: $(\theta_t^{base}, \{\theta_t^{arm}\}) \leftarrow (\hat{\theta}_t^{base}, \{\hat{\theta}_t^{arm}\})$\;
        }
        Update annealing temperature $T_{anneal} \leftarrow \text{cool}(T_{anneal})$\;
    }
}
\Return{$\{\theta_t^{base}, \{\theta_t^{arm}\}\}_{t=0}^{T}$}
\end{algorithm}

\begin{figure}[t]
  \centering
  \setlength{\abovecaptionskip}{-0.05cm}
  \includegraphics[width=1.0\linewidth]{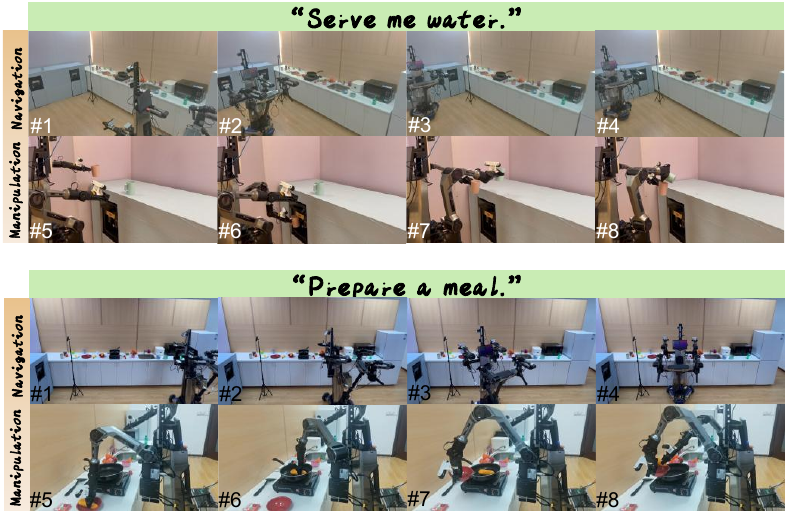}
  \caption{Visualization of mobile manipulation trajectories for real-world experiments.}
  \label{fig:manipluation_vis}
  \vspace{-0.8cm}
\end{figure}

\begin{figure}[t]
  \centering
  \setlength{\abovecaptionskip}{-0.05cm}
  \includegraphics[width=1.0\linewidth]{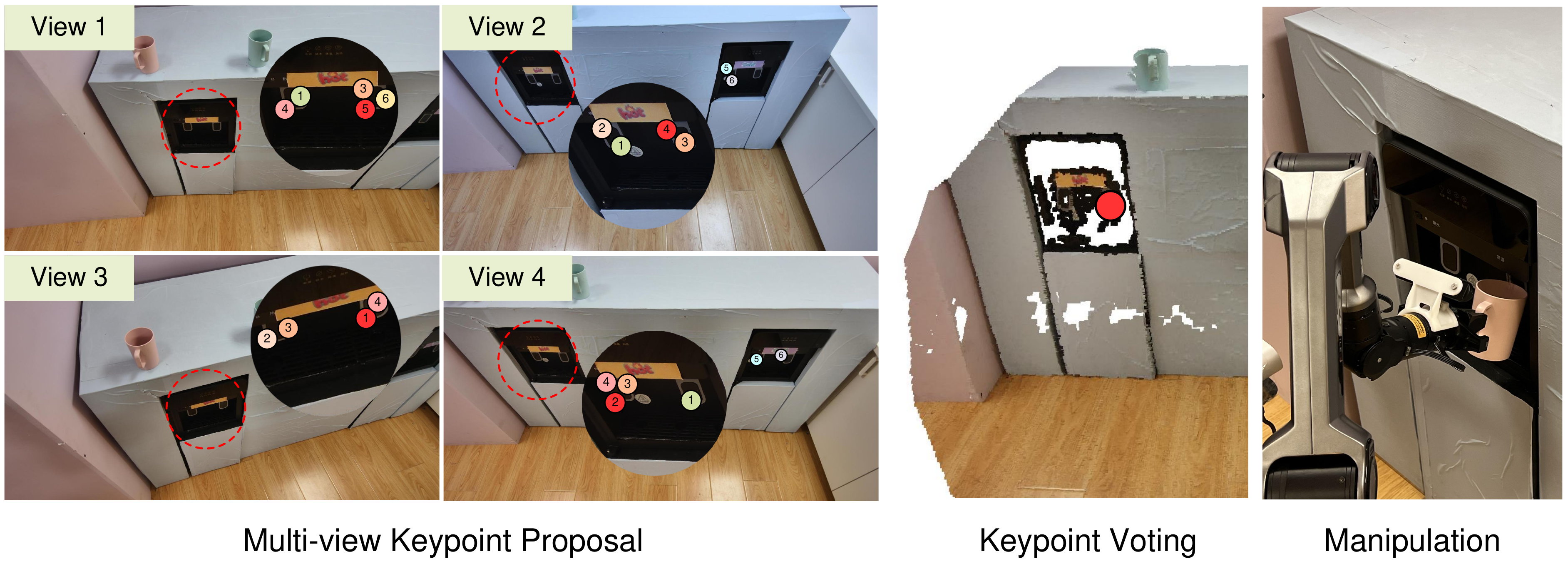}
  \caption{Visualization results for keypoint generation. MoTo selects keypoint proposals (red points) from multi-views, projects them into 3D space and votes to generate keypoints for manipulation.}
  \label{fig:keypoint_vis}
    \vspace{-0.2cm}
\end{figure}

\begin{figure}[t]
  \centering
  \setlength{\abovecaptionskip}{-0.1cm}
  \includegraphics[width=1.0\linewidth]{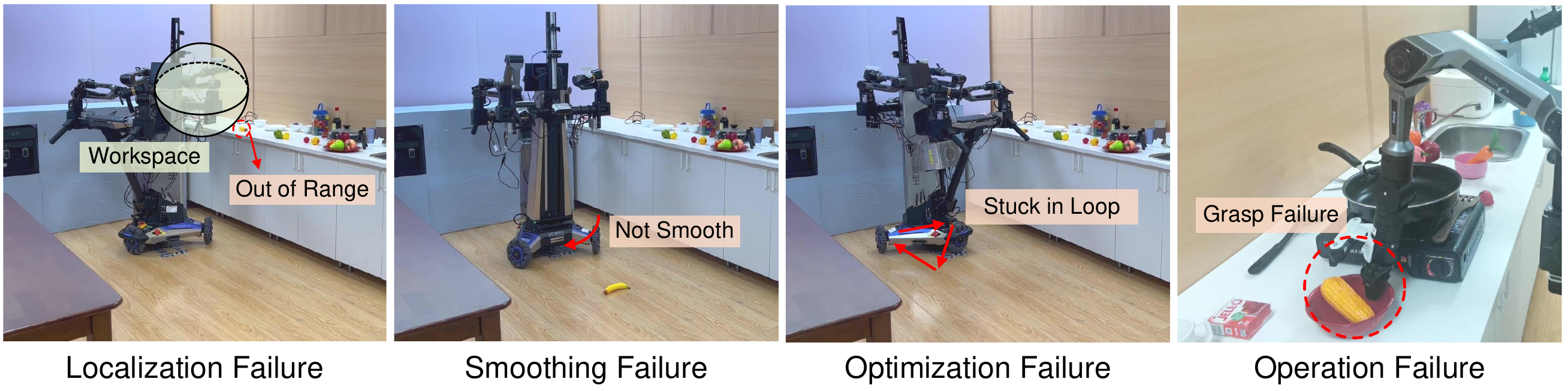}
  \caption{Failure Cases in real-world experiments.}
  \label{fig:failure_case}
  \vspace{-0.5cm}
\end{figure}

\section{More Results}
In this section, we present additional experimental results. 
We visualize mobile manipulation trajectories for tasks “Serve me water” and “Prepare a meal” in Figure~\ref{fig:manipluation_vis} to demonstrate the performance of MoTo on mobile manipulation.
We further show failure results to identify the limitations of the existing work and better help relevant researchers to follow this work.
\subsection{Manipulation Visualization}
Figure~\ref{fig:keypoint_vis} demonstrates the scene keypoint generation and mobile trajectory in task “Serve me water”. MoTo first extracts keypoint candidates from multi-view RGB images with VLM, and then projects them into the 3D space to vote for specific regions to be interacted with utilizing the consistency of the viewpoints. Based on the key point guidelines, MoTo searches for mobile trajectories using a zero-shot optimization algorithm, and generates feasible docking points for subsequent action, to execute the subsequent fixed-base manipulation.


\subsection{Failure Cases}
Figure \ref{fig:failure_case} illustrates 4 common failures in the experiment, which are analyzed as follows:

\textbf{Smoothing Failure:} In real-world experiments, the unsmoothed motion trajectory of the base results in high acceleration, triggering the protection mechanism, which leads to the failure of mobile manipulation. The smoothing failure in Figure \ref{fig:failure_case} illustrates that the neighboring waypoints on the mobile manipulation trajectory are distant from each other, and the excessive acceleration of the mobile base activates the protection mechanism, causing the banana to fall.

\textbf{Localization Failure:} SLAM is the key for the robot to reconstruct the scene, and also provides real-time position information to the robot. Errors due to SLAM can cause the robot to stay in a position far away from the specific target object.

\textbf{Optimization Failure:} Using the Dual Annealing algorithm sometimes gets stuck in a loop, causing the robot to keep adjusting the base position repeatedly.

\textbf{Operation Failure:} Due to a series of errors such as SLAM, the robot stops at a different position each time, which presents a challenge to the viewpoint generalization of the fixed-base manipulation model. 
How to improve the viewpoint generalization of the manipulation policy is one of the key bottlenecks in mobile manipulation.

\end{document}